\documentclass[10pt, logo, copyright]{scafo}
\usepackage{csquotes}
\usepackage{cleveref}

\usepackage{mdframed}
\usepackage{listings}

\AtEveryBibitem{
\ifentrytype{inproceedings}{
 	\clearlist{address}
 	\clearlist{publisher}
 	\clearname{editor}
 	\clearlist{organization}
 	\clearfield{url}  
 	\clearfield{doi}  
 	\clearfield{pages}  
 	\clearlist{location}
 }{}
 \ifentrytype{article}{
	 \clearfield{url}
 }
 }
\newcommand{\xxnote}[3]{}
\ifx\hidenotes\undefined
  \usepackage{color}
  \renewcommand{\xxnote}[3]{\color{#2}{#1: #3}}
\fi

\usepackage{soul}
\newcommand{\method}[1]{GRID}
\def\code#1{\texttt{#1}}
\usepackage[dvipsnames]{xcolor}
\definecolor{llmbackground}{RGB}{212, 233, 239}
\def\user#1{{\color{Gray}#1}}

\def\breaktext#1{\vspace{1mm}\\}

\usepackage[skins]{tcolorbox}
\def\llm#1{\colorbox[RGB]{212, 233, 239}{\parbox{0.98\textwidth}{GRID> \ #1}}}
\def\codebox#1{\colorbox[RGB]{242,242,234}{\parbox{0.9\textwidth}{\code{{\color{Black}#1}}}}}

\usepackage{listings}
\definecolor{codegreen}{rgb}{0,0.6,0}
\definecolor{codegray}{rgb}{0.5,0.5,0.5}
\definecolor{codepurple}{rgb}{0.58,0,0.82}
\definecolor{backcolour}{rgb}{0.95,0.95,0.92}

\lstdefinestyle{mystyle}{
  backgroundcolor=\color{llmbackground}, commentstyle=\color{codegreen},
  keywordstyle=\color{magenta},
  numberstyle=\tiny\color{codegray},
  stringstyle=\color{codepurple},
  basicstyle=\ttfamily\scriptsize,
  breakatwhitespace=false,         
  breaklines=true,                 
  captionpos=b,                    
  keepspaces=true,                 
  numbers=none,                    
  numbersep=5pt,                  
  showspaces=false,                
  showstringspaces=false,
  showtabs=false,                  
  tabsize=2,
  xleftmargin=10pt,
  framesep=8pt,
  frame=l,
  columns=fullflexible
}

\usepackage{inconsolata}

\usepackage[export]{adjustbox}

\newcommand\rurl[1]{%
  \href{http://#1}{\nolinkurl{#1}}%
}

\title{\method{}: A Platform for General Robot Intelligence Development}
\author{Sai Vemprala, Shuhang Chen, Abhinav Shukla, Dinesh Narayanan and Ashish Kapoor \\ \href{https://scaledfoundations.ai/}{\bf{Scaled Foundations}}}

\begin{abstract}

Developing machine intelligence abilities in robots and autonomous systems is an expensive and time consuming process. Existing solutions are tailored to specific applications and are harder to generalize. Furthermore, scarcity of training data adds a layer of complexity in deploying deep machine learning models. We present a new platform for General Robot Intelligence Development (\method{}) to address both of these issues. The platform enables robots to learn, compose and adapt skills to their physical capabilities, environmental constraints and goals. The platform addresses AI problems in robotics via foundation models that know the physical world. \method{} is designed from the ground up to be extensible to accommodate new types of robots, vehicles, hardware platforms and software protocols. In addition, the modular design enables various deep ML components and existing foundation models to be easily usable in a wider variety of robot-centric problems. We demonstrate the platform in various aerial robotics scenarios and demonstrate how the platform dramatically accelerates development of machine intelligent robots.The GRID platform can be accessed at \url{https://github.com/ScaledFoundations/GRID-playground}.

\end{abstract}

\begin{document}
\maketitle

\section{Introduction}
Without the ability to sense, reason, and act appropriately, robots and autonomous systems are nothing more than a collection of expensive parts. The inability to leverage machine intelligence to deliver autonomous systems with these capabilities is arguably the most substantial hindrance to adoption. At the core, the challenge facing artificial intelligence (AI) in robotics involves solving the perception-action loop; in which, provided a sensory motor stimulus the machine needs to decide on the next action to take. In a robotic system there are often many such perception-action loops running in parallel or sequentially, and sometimes also embedded hierarchically

There are several reasons for the modest impact of AI within robotics. First, too often solutions are tailored to one specific application or problem, leading to idiosyncratic implementations that fail to generalize. 
Second, it is common for techniques and components to be stitched together to provide a tailor made solution that is compatible or optimized for the sensor suite on the robot and the environment it will operate in. 
Custom solutions are expensive to engineer and are not amenable to adaptation when the task configuration changes. Third, the available tools are not well suited to robotics. For example, there is a dearth of foundational AI models for robots and a lack of good tooling to enable tuning and real-world deployment.

To address this limitations head on we propose a solution: {\em General Robot Intelligence} which we define as the ability for a robot to learn, compose, and adapt skills to their physical capabilities, environmental constraints and goals. For example, a collision avoidance mechanism in one kind of robot should be useful for another robot for a similar purpose even if the sensor configurations differ. 
In order to learn a novel skill unique to the robot, the machine should be able to re-purpose existing machine intelligence modules instead of starting from scratch. 
A robot intelligence methodology that offers modularity, scalability and generalizability across a wide variety of robots, sensor suites and environments will enable rapid implementation of machine intelligence on robots and lower the barrier to entry.

Many fields within AI (e.g., Computer Vision, Natural Language Processing, Acoustics and Speech Processing) had a similar phase where researchers and engineers produced point solutions that failed to generalize beyond a specific use case. However, the state-of-the-art methodologies in all of these fields now do not start from scratch but instead builds upon the core idea of Foundation Models. Foundation Models are large Neural Networks that are pre-trained on large amounts of data and are used as a starting point for almost all of the tasks in a given domain. Examples of such Foundation Models include GPT\cite{brown2020GPT-3}, LLaMA\cite{touvron2023llama}, Gato\cite{gato}, RT-2\cite{rt22023arxiv}, Segment Anything\cite{kirillov2023segany}, ImageBind\cite{girdhar2023imagebind} etc.

A Foundation Model for robotics is central to our efforts in achieving General Robot Intelligence. However, there are aspects of robotics that make building Foundation Models different than other sub-fields of AI. First, unlike other domains the data available for pre-training such Foundation Model is extremely scarce. For instance, there is no webscale data that can be crawled or scraped. At the same time, collecting data from real-world robots is expensive and challenging. Secondly, the nature of robot telemetry is multimodal due to a variety of sensors that are used. It is noteworthy that the multimodal data extends beyond just the combination of language and images, which traditionally has been focus of the most of the recent efforts to build multimodal Foundation Models. The heterogeneity of sensors, their placement and varied sampling frequencies are challenges that are unique to robotics and are not considered in a majority of approaches. Third, it is far important in robotics to have Foundation Models that are compatible with edge processing. Moreover, we expect many different perception-loops that need to be implemented on a robot thus requiring much more computationally efficient modeling. Finally, the robots are devices that take actions in the real-world and an incorrect decision can do a lot of harm. Embedding and enabling safe behavior of robots via Foundation Models is a question that is of paramount importance. 

This paper outlines our strategic road map and progress towards building a General Robot Intelligence Development (\method{}) platform. At its core, the platform features Foundation Models engineered to be the catalysts for a myriad of AI-centric robotic tasks. These models span a wide range, catering to specialized functions like perception, state estimation, mapping, planning, control - all of which are core components in any robot autonomy pipeline. Foundation Models are a natural step toward General Robot Intelligence as these large models have the ability to capture not only the structure of the physical world, but also learn how such physical reality is perceived via the different sensors. With appropriate design and enough data the models also capture the dynamics of the system and can understand the causal connection between action and change in the system state. 

To tackle the challenge of data scarcity, we introduce the concept of {\em Foundation Mosaic}, where several existing foundation models from multiple domains are synergistically incorporated to reduce the sample complexity required for pre-training Robotics Foundation Models. Our framework marries state-of-the-art large-scale simulations with this ensemble of foundation models, which further help with scalable training data generation as well as evaluation. We also discuss novel methodologies, such as Simulation Feedback, to refine and augment robotic intelligence capabilities. Furthermore, we detail a multi-tiered approach to safety, extending from licensing protocols to the intrinsic architecture of Foundation Models that prioritize secure and reliable operation. 

Through the GRID platform, we seek to kickstart a new way of building robots, one that supersedes the limitations of application-specific intelligence and allows a more versatile, general robot intelligence.
\section{GRID}
\method{} aims to democratize robotics, making it more accessible to researchers, developers, and organizations alike. Traditional barriers to entry, such as the requirement for specialized knowledge in various sub-disciplines and the high cost associated with data collection and model training, have deterred many from making contributions. \method{} addresses these challenges by providing a comprehensive, modular architecture that integrates rich multimodal data, foundation models, and large language models (LLMs). Drawing inspiration from how models like GPT-4 and Copilot have helped accelerate software development, \method{} seeks to be the equivalent catalyst in the realm of robotics. By offering an all-inclusive, user-friendly framework, GRID hopes to accelerate innovation and broaden participation in the development of intelligent, adaptable, and responsible robotic systems.

\begin{figure}
    \centering
    \includegraphics[width=\linewidth]{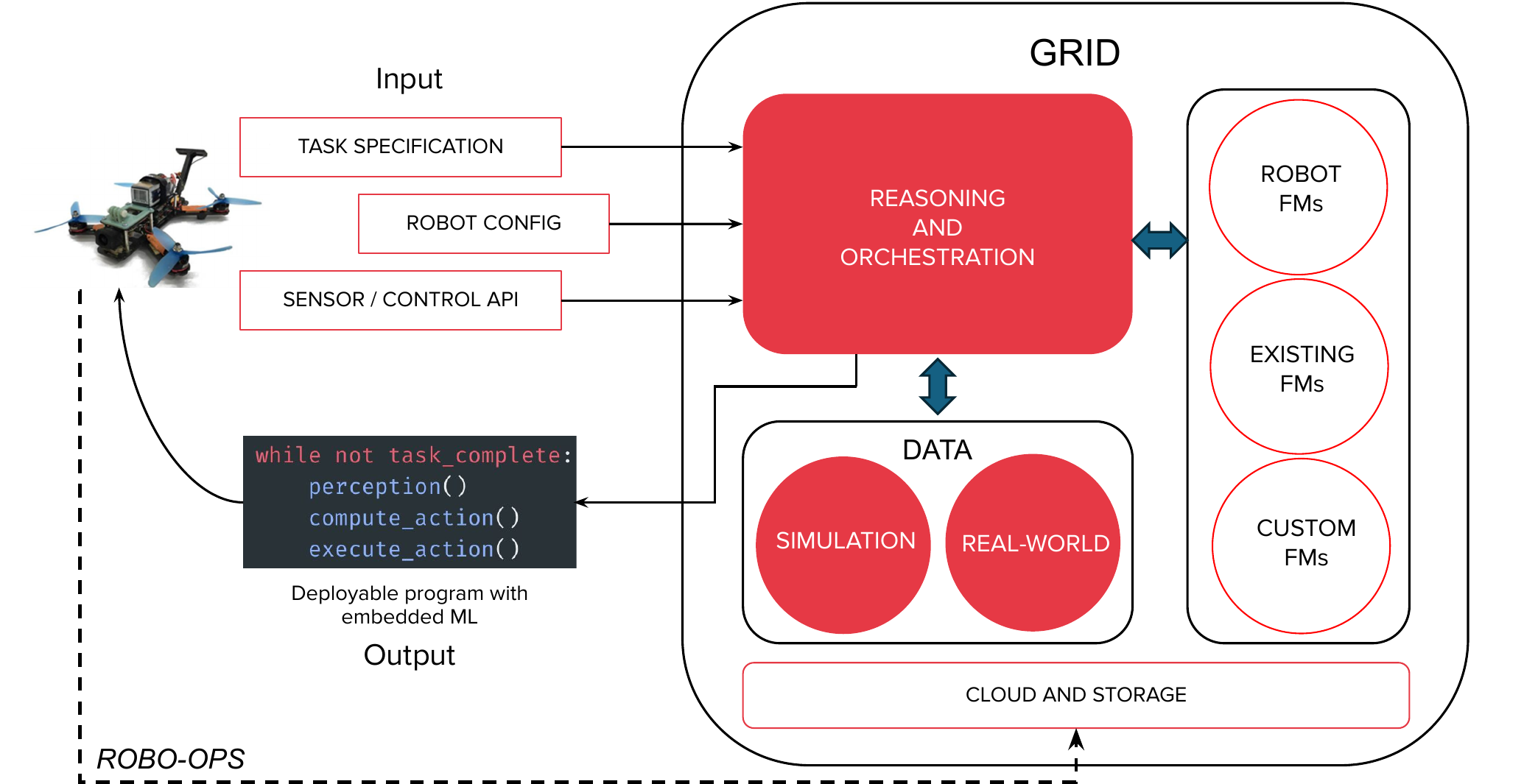}
    \caption{An overview of the \method{} platform. \method{} can orchestrate a set of Foundation Models and other tools to solve complex robotics tasks using both simulation and real-world data. The orchestrator uses APIs to interact with the robot, world, and models, to produce a deployable program with embedded ML that solves the specified task.}
    \label{fig:grid_overview}
\end{figure}

\subsection{Model Architecture}
Figure \ref{fig:grid_overview} illustrates the core components of the \method{} framework. Given robot configuration, task specification and details about the sensor suite and robot control APIs, \method{} generates code with relevant ML modules to achieve the goal. Central to \method{} are the robot foundation models acting as the core of the machine intelligence, which will enable the machine to make non-myopic decisions under uncertainty and incomplete information. The second component involves our ability to pre-train such large neural networks - data scarcity in robotic domains makes it non-trivial, consequently our architecture builds upon both high-fidelity simulations as well a framework to curate real-world data. Additionally, instead of pre-training from scratch, we build the foundation models via mosaicing many existing foundation models that are pertinent to robot perception and action. Access to high-fidelity simulations also means that our ability to pre-train, perform inference and fine-tuning can benefit from novel learning mechanisms such as simulation feedback. Finally, safe and efficient deployment of robot intelligence in the real-world is of paramount importance and \method{} seeks to build upon advances in these areas. Next, we describe each of the components in detail.

\subsubsection{Foundation Mosaics}
The field of artificial intelligence has undergone a significant shift over the past few years, moving from an array of task-specific, narrow models to larger, generalized neural networks. For instance, in the area of natural language processing (NLP), models like GPT-4 have exhibited a remarkable range of capabilities, including text summarization, translation, and sentiment analysis. In parallel, visual-language models are becoming increasingly proficient in multiple tasks such as object detection, image captioning, and even generative tasks like artwork creation.

This development suggests that a single, generalized model could feasibly replace a multitude of task-specific models, leading to increased efficiency and more straightforward system architecture. However, when transitioning from domains like NLP to robotics, several complexities arise. First, there is a scarcity of data - Foundation Models require large amounts of training and there are limited curated datasets for robot pre-training. Consequently, it is far important to build intelligence capability over existing Foundation Models, even when their intended use case was different than robotics. For example, the Foundation Models for perception \cite{kirillov2023segany, yolov8} can be useful for various scenarios. Similarly, recent work on pre-training for decision-making can be useful \cite{gato, yanchao_iclr}. 

Additionally, in robotics the variability in actuators and control systems adds an extra layer of complexity. Each robot type—be it a quadcopter, an industrial manipulator, or a mobile rover—has a unique set of actuators and corresponding control systems. Therefore, the notion of generalization becomes trickier. Using a monolithic neural network that maps sensor inputs directly to actuator outputs is not scalable, and also runs the risk of not capitalizing on years of existing research into control theory. In many mobile robotic form factors such as wheeled robots or aerial vehicles, robust control algorithms already exist for skills like trajectory tracking.

\method{} uses an ensemble approach that we term \emph{Foundation Mosaic}, which is composed of individual foundation models. The key insight is that robot intelligence requires a collective composition of expertise. 
Instead of operating in isolation, the models are orchestrated through a Large Language Model (LLM)-based fusion mechanism that continually integrates their outputs into coherent structures representing environment semantics, physics, and language grounding. 

\begin{figure}
    \centering
    \includegraphics[width=\linewidth]{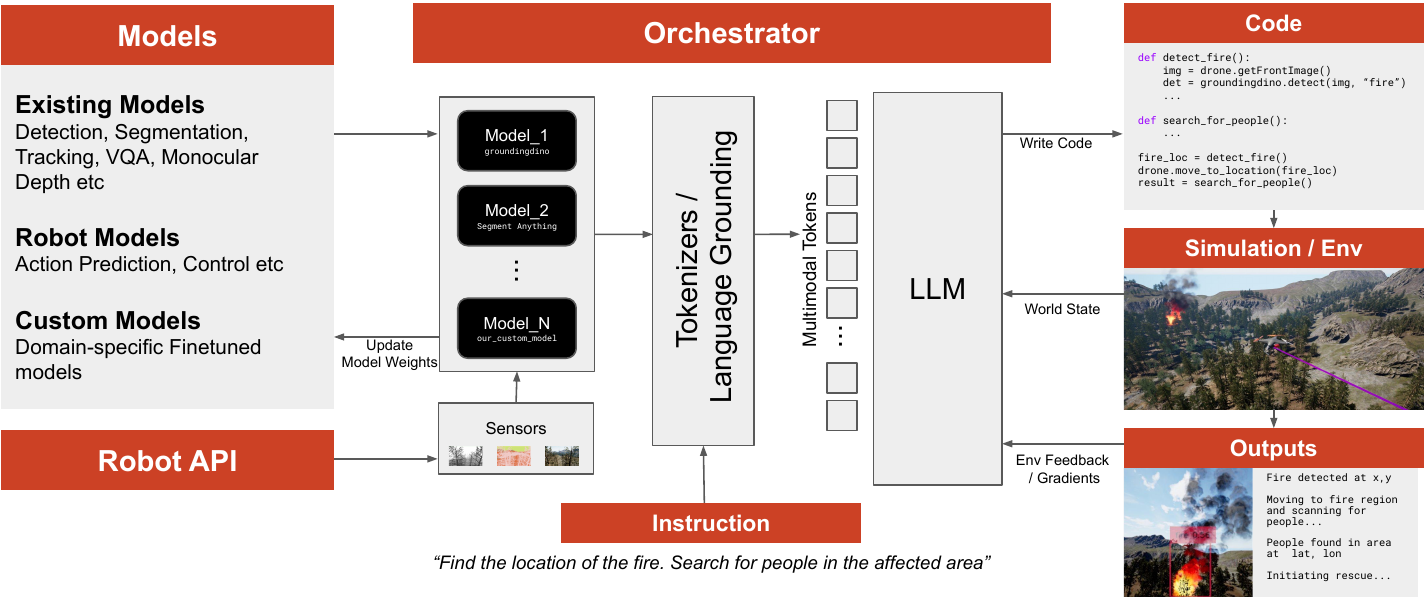}
    \caption{An illustration of the Foundation Mosaic. The Orchestrator uses an LLM to intelligently compose a set of available models and the sensors of the robot operating in the environment to solve the specified task. The models produce a language-grounded representation (which could be either a text-based state summary or features/tokens in a shared multimodal representation space). The LLM uses this information in a feedback loop to write and execute code that makes the robot take actions in the environment. The crux of the Foundation Mosaic is the ability to seamlessly leverage domain-specific information from large-scale Foundation Models while creating an understanding of the environment that is grounded in language.}
    \label{fig:foundation_mosaic}
\end{figure}

\Cref{fig:foundation_mosaic} shows the architecture of the Foundation Mosaic, where existing Foundation Models are building blocks that plug into the decision-making framework and interface naturally with control primitives. Instead of building everything from scratch, such a modular approach allows for better specialization while retaining the benefits of large-scale foundation models. For instance, object detection and instance segmentation, tasks traditionally covered by specialized computer vision models, can still form part of a larger system. Their outputs are fused with other data streams to create a unified representation of the robot's operational environment. This collective output would be rich in both geometric and semantic features, thereby offering the robot a more comprehensive understanding of its surroundings.

Note that the Foundation Mosaic is a neural architecture. Consequently, the numerical procedures that train a neural network enable fine-tuning of the entire model or parts of the model that might require additional refinement. This is an appealing property as now the architecture has the ability to retain specialty while assembling a breadth of skills that can be performed.
While an ensemble of Foundation Models are the core ingredients, they need appropriate orchestration, fusing and scaffolding so that the different representations can work together. Such orchestration and unification is non-trivial which \method{} accomplishes with the reasoning capability of large language models.
    
\subsubsection{Reasoning with Large Language Models }
The remarkable capabilities of large language models like GPT-4 have set a new standard in complex reasoning, contextual understanding, and symbolic inference. These models have demonstrated unparalleled expertise in a range of natural language processing tasks, from text generation and question-answering to sentiment analysis. Similarly, vision-language hybrid models have emerged as an exciting area of research aiming to unify visual and textual data. 

\method{} uses LLMs as orchestrators in the foundation mosaic of specialized models. Most commonly given documented capabilities of individual models (neural or otherwise), the orchestrator first can select relevant modules. The LLMs can then synthesize the outputs of the selected modules into actionable insights or even deployable code, offering a highly efficient mechanism for human-machine interaction. Not only can users converse naturally with their robots, but they can also provide immediate feedback, which is instrumental for the model's continuous learning and adaptability. 

Within the language model layer of GRID, we utilize a multi-agent framework where two LLMs play the roles of \emph{Actor} and \emph{Critic}. When the human user commands a specific task, the actor responds to with a step-by-step plan, which can be evaluated by the critic. Any modifications suggested by the critic are piped back to the actor, and this cycle continues until the plan is optimal (human feedback can be sought out at any stage). Once the plan is final, the actor outputs code which is checked for correctness, and any execution output is again sent back to the actor for necessary corrections. Finally, environmental feedback from the simulation (e.g. did the robot do what it was supposed to do?) is also monitored and sent back to the actor for further refinement. This structure follows the growing consensus that multi-agent debate or multi-round verification improves the reasoning power and correctness of LLMs \cite{du2023improving}, and we have found empirically that it reduces the potential for hallucinations or errors.

Also, the individual models that compose the mosaic can also be considered as informative features that can plug into the LLM layers via a tokenization mechanisms. Intuitively, the Foundation Mosaic generates rich multimodal representations, which are then grounded via an LLM and thus enable reasoning required to orchestrate and fuse outputs to accomplish the eventual robotic task. For example, in an aerial navigation scenario, the LLM can use the outputs of a vision model to generate a textual summary of the robot's environment, figure out that the sky is getting cloudy, query a weather model for more information, reason about it, and then generate a plan of action.

Note that such modeling of perception-action loops via an LLM-powered Foundation Mosaic makes language an integral part of robot intelligence. Furthermore, the LLMs not only serve as reasoning agents but also as intuitive interfaces for human-machine interaction. They allow users to engage in natural conversations with robots and offers an inherent layer of explainability that is essential for regulatory compliance and operational analysis.

\subsection{Data: Multimodality and Scarcity}
The complexities inherent in robotics stretch far beyond those of the majority of current machine learning applications that predominantly focus on the notion that multimodality is limited to visual and textual data. The richness of sensory data types necessary for effective robotic systems places a premium on the multimodal nature of data. These encompass not just conventional RGB images and audio signals, but also more specialized data forms such as haptic feedback, depth maps, temperature readings, and LiDAR point clouds. Each of these modalities contributes unique informational facets that are essential for building a robot capable of complex reasoning, planning, and real-time adaptability to varying environmental conditions.

However, the integration of these diverse data types introduces its own set of challenges. One such issue pertains to the temporal alignment of data. For instance, sensors like RGB cameras typically collect data at a much lower frequency than Inertial Measurement Units (IMUs), which capture motion and orientation data at higher speeds. Addressing the discrepancies in data rates requires robust techniques for time-sensitive data alignment, interpolation, and feature fusion, without which the resulting robotic models could suffer from increased latency, reduced robustness, and operational inconsistencies.

Beyond the complexities of multimodal data integration, the domain of robotics also faces a more foundational obstacle: the scarcity of rich, large-scale datasets. In contrast to areas like natural language processing and computer vision, where massive, annotated datasets exist, robotics frequently encounters constraints in data collection. Often, this data collection process is not only resource-intensive but also confined to highly specialized and constrained environments, making it inherently limited in scope and applicability.

\subsubsection{Robot Simulation for Multimodal Data Generation}
Given the constraints and complexities related to multimodal data, we posit that simulation is an essential component in the development of general robot intelligence. A key part of the GRID framework is a high fidelity simulation platform named AirGen. AirGen is built upon Microsoft AirSim \cite{shah2018airsim}, and focuses on aerial robotics scenarios. AirGen is designed to be versatile, capable of emulating a wide array of both synthetic and real-world, geo-specific environments. This is fundamental for generating a sufficiently diverse set of scenarios that a robot may encounter, thereby significantly improving model robustness and adaptability. One of the core features of this simulation environment is its ability to produce spatio-temporal traces of trajectory data that can be virtually infinite in size. For example, in an aerial robotics context, several variables like time of day, lighting conditions, trajectory dynamics, altitude levels, and regional specifics can be randomized to provide a comprehensive set of training scenarios. Such domain randomization techniques extend the utility of simulation beyond mere data generation to creating more generalized and robust robotic models. AirGen contains integrations with geospatial data sources such as Bing Maps and Google Maps, as well as high-precision terrain data from several sources.

In light of recent advancements in simulation technology, the potential for generating even more realistic and diverse data has expanded considerably. For instance, neural scene reconstruction methods \cite{mildenhall2021nerf, muller2022instant, kerbl20233d} offer a way to construct highly detailed and photorealistic 3D environments, effectively serving as digital twins of real-world scenarios. Such digital twins provide a high-fidelity training ground for robots, particularly for tasks that demand a strong resemblance to actual environmental conditions. Generative AI technologies also present promising possibilities. Unlike classical rendering techniques, which may have limitations in terms of realism and diversity, generative models can synthesize entirely novel, yet plausible, environmental conditions and scenarios from textual descriptions or other priming. Such capabilities offer a way to expand the data distribution, thereby introducing elements of unpredictability and stochasticity that better mirror the complexities and uncertainties of real-world conditions. AirGen contains API for importing meshes at runtime, which allows users to bring in their own objects (or potentially even scenes) to simulate scenarios of interest. The ability to generate multimodal data from a variety of scenes also allows for the training of generative models that can be used to augment classically rendered or real-world data.

In conclusion, the role of the simulation platform in the context of GRID's approach to robotics is twofold. First, it addresses the pressing issue of data scarcity by providing a mechanism for generating rich, multimodal data at scale. Second, the advancements in simulation technologies, including the integration of neural rendering and generative AI, pave the way for creating data of unparalleled quality and realism. Collectively, these features contribute to the overarching goal of developing robotic systems that are not merely functional but are also highly intelligent, adaptive, and capable of interacting coherently and meaningfully with their complex environments.

\subsubsection{Real-world}
While significant volumes of data from real-world robotic operations are generated today, most organizations deploying robots for specific use cases are not considering benefits to structuring, aggregating, and sharing this data for global model training, beyond optimizing current performance of their own systems.   

In most cases, the effort to collate real-world robotic data is significant and requires process considerations, including building ingestion pipelines from multiple data sources and types, associating and normalizing this data given challenges referenced above, and introducing specific steps in scale operations of these robots to extract or in some cases manually transfer data from robot to data estate.   

Additionally, human robot operators completing similar tasks with robots have inherent variances on operating safety tolerances and ways they choose to complete tasks with a specific robot type.  For example, manual drone operators inspecting an energy asset, such as a wind turbine, may fly drones at different speeds, at different proximities to the asset to facilitate data collection, and at different trajectories from origin to destination based on skill level and experience.   

For a foundation model training data set to be truly comprehensive and support optimal AI-powered tasks completion, we posit that this data set should reflect as many operating permutations and combinations of multi-modal data as possible, across a broad set of operating environments and conditions.  This critical need for data diversity further bolsters the need to supplement real-world data with high quality, imperceptibly accurate synthetic data.   

Beyond effort, process, and operating variance, additional technical considerations include point of data processing within a robotic architectural deployment (on board the robotic “edge” vs. at a local gateway or within a cloud computing back-end) and existence and persistence of connectivity to the robot end point.

\subsection{Learning, Inference and Fine-Tuning}
As robotics aims to bridge the gap between controlled laboratory settings and the unpredictability of real-world applications, the chaotic dynamics of real-world environments can present unforeseen challenges. Simulated environments, therefore, provide a controlled space where robots can iteratively refine behaviors, reduce errors, and align with desired outcomes without real-world risks. Simulation provides full information of robot's perception-action loop that enables step-wise evaluation and feedback. Within the \method{} framework, the step-wise feedback granted by simulation is key to achieving complex reasoning required for high-level robotic intelligence. 

Learning from feedback, particularly within a simulated environment, offers a multitude of benefits and can be implemented in several innovative ways. For example, simulation feedback plays a pivotal role in ensuring that the developed robotic systems are both competent and safe. Below, we explain several ways simulation feedback can be used within \method{}:

\emph{Gradient-free Learning from Simulation Feedback:}
This approach sees the LLM tasked with solving a specific robotics challenge, such as obstacle navigation or object detection. \method{} initiates this by writing code that leverages a collection of models, each honed for specialized tasks. However, the generated code might not always be perfect or optimal. By deploying this code within a simulation environment, we obtain real-time feedback regarding its performance.

The simulator not only provides metrics that indicate success or failure but can also supplement this with detailed ground truth data and other relevant feedback. With this feedback in hand, the LLM can iteratively refine its code, with the simulation playing a continuous role in evaluation. This back-and-forth continues until the LLM produces code that effectively solves the given task within the simulation constraints.

\emph{Reinforcement Learning from Simulation Feedback (RLSF):}
This method offers a more involved approach where the LLM is fine-tuned using feedback derived directly from the simulation. Much like how ChatGPT was enhanced via human feedback, the LLM leverages simulation feedback to improve its performance and alignment with the task objectives. The core advantage here lies in the realm of safety and alignment. When we consider robotics, these two aspects become paramount, especially when deploying robots in real-world scenarios. By finetuning the LLM in a controlled simulated environment, we can ensure that it behaves predictably and in alignment with safety protocols once deployed in the real world.

\emph{Fine-tuning Individual Foundation Models:}
At times, while the LLM can orchestrate tasks efficiently, the foundation models, or the "mosaic", might not perfectly align with the domain of interest. For instance, aerial robotics introduces challenges like recognizing distant objects or adapting to camera orientations that are not common in conventional vision models.

In scenarios like these, there's an evident distribution gap between the inherent representations within the foundation models and the specialized requirements of the domain. One way to bridge this gap is to employ simulation feedback to directly fine-tune the individual foundation models. By deploying these models within simulations that mimic specific challenges, we can generate domain-specific feedback. This feedback, often supplemented with simulation-generated labels, can then be used to backpropagate gradients and refine the models, ensuring that they are better suited for the specific robotics task at hand.

With massive datasets gathered from simulation, the Foundation Mosaic can be trained as a whole in an end-to-end fashion. Its individual parts can also be targeted and fine-tuned directly for specific scenarios. Most importantly, we believe the step-wise feedback granted by simulation is the key to achieving complex reasoning need for high-level robotic intelligence.

ChatGPT for Robotics \cite{vemprala2023chatgpt} showcased basic reasoning capabilities from large language models (LLM) as an orchestration layer for robotics applications. However, the reasoning capability is still considered as emergent property via prompting LLMs that are originally fine-tuned on single-round datset (output desired response given prompt) \cite{ouyang2022training, wei2023chainofthought}. High-level robotic intelligence encompassing long-horizon planning \cite{yao2023tree, yang2023large}, safety precautions, and failure recovery demands much more sophisticated multi-step reasoning. \cite{lightman2023let} showed training on step-level human feedback greatly improves LLM's complex multi-step reasoning in solving math problems. It is also well-known in reinforcement learning that dense step-level rewards improves sample complexity \cite{ng1999policy}.

\subsection{Safety}
Safety is a very important consideration in robotics and autonomous systems. Introducing an element of AI in robotics exacerbates the problems due to issues of partial observability and difficulties in characterizing the behavior of the neural networks. While there is a wide-spectrum of discussion on AI safety, here we primarily discuss safety issues due to bugs or incompleteness in AI engineering. There are multiple communities who have been tackling this difficult problem and \method{} can be a useful platform for furthering those ideas. Examples include work on Safety Barrier Functions \cite{ames2019control,luo2022sample, castañeda2022probabilistic}, Formal-methods based approach to safety \cite{reactiveSTL,sertac,Sadigh2016SafeCU} , safe reinforcement learning that have used robots and robot simulators to further those ideas.

Robot safety in the realm of foundation models, however, is a less explored topic. Note that strong reliance on a Neural Network trained from scratch can be problematic due to their vulnerability to overconfident or unpredictable results when presented with data distributions unseen during training. Additionally, in real-life scenarios there might be environmental factors and uncertainties due to machine perception, that may not have been explicitly modeled in the formulation. Foundation models for robots trained on a large amount of training data might circumvent this problem. However, this stays an important open problem as we bootstrap bigger robot foundation models with large-scale data.

We'd also like to highlight that \method{} along with its foundation models provides an excellent playground to test and try new ideas before deploying them in the real-world. Additionally, mechanisms such as simulation feedback and generative data capabilities equips researchers and engineers to understand corner cases and failure modes \cite{shromona}.

Given the nascency of the field and lack of definitive algorithmic solutions, it is fairly important that the researchers and engineers take a responsible approach towards building and deploying the solutions. \method{} specifically depends upon mechanisms such as Responsible AI Licenses (RAIL) \cite{RAIL} to vector efforts of the community in a direction that is safe and beneficial for humanity. 
RAIL were developed as part of a community initiative in response to the growing demand for more concrete tool to help practitioners address AI misuse. OpenRAIL licenses foster openness of technology (including source code, models, and apps) but with behavioral use terms that address unethical use of the technology. As an example, behavioral use licenses are now backed by some of the largest open source generative AI projects to extend access while limiting negative outcomes.

\subsection{Deployment on Robots: Foundation Models on the Edge}

Foundation Models are often trained in a large-scale distributed manner on expensive and high performance hardware. The best models trained on web-scale data usually cost millions of dollars to train and the GPUs they run on have hundreds (or thousands) of teraflops of compute power. In contrast, the actual devices / robot form-factors that the models need to run on are severely constrained in terms of on-device compute, and bring a number of other challenges that range from spotty (or non-existent) internet connectivity in remote deployment scenarios to specific edge hardware considerations like memory, power, and latency.

Model optimization techniques like distillation, quantization, pruning, and neural architecture search are all useful techniques to reduce the computational requirements (flops, parameter counts, inference latency) of running a model on an edge device while trying to optimize for the specific deployment hardware. There has been a lot of recent work on retraining and deploying equally performant models with lower-precision number formats (fp16, bf16, fp8) and integer-quantized (e.g. int8 and int4) weights \cite{dettmers2023case, dettmers2022llm}. Techniques like FlashAttention \cite{dao2022flashattention} leverage the hardware's characteristics (like memory hierarchy) to deliver further performance gains. A combination of these techniques can dramatically reduce the inference footprint of a Foundation Model and make it possible to deploy such a model on a robot form factor.

Another related line of work relevant to \method{} is parameter-efficient finetuning techniques for Foundation Models. Methods like QLoRA \cite{dettmers2023qlora} and Prompt Tuning \cite{lester2021power} enable a smaller set of additional parameters to be finetuned while keeping the base Foundation Model frozen. QLoRA has even made it possible to finetune models with billions of parameters on inexpensive laptops.

We anticipate \method{} being able to use a combination of these techniques to facilitate efficient finetuning and domain-specific adaptation of Foundation Models on robot form factors.

In scenarios with stable internet connectivity, there is also the possibility of hybrid cloud + edge deployment. A more efficient but less performant model can be deployed on the device while a larger and more performant model deployed on the cloud can intermittently run inference and guide/denoise the predictions of the smaller edge model. \method{} can support such a hybrid deployment out of the box by adding the cloud and edge model implementations to the Foundation Mosaic.

\section{Preliminary Results and Examples}

\begin{figure}
    \centering
    \includegraphics[width=1.0\columnwidth]{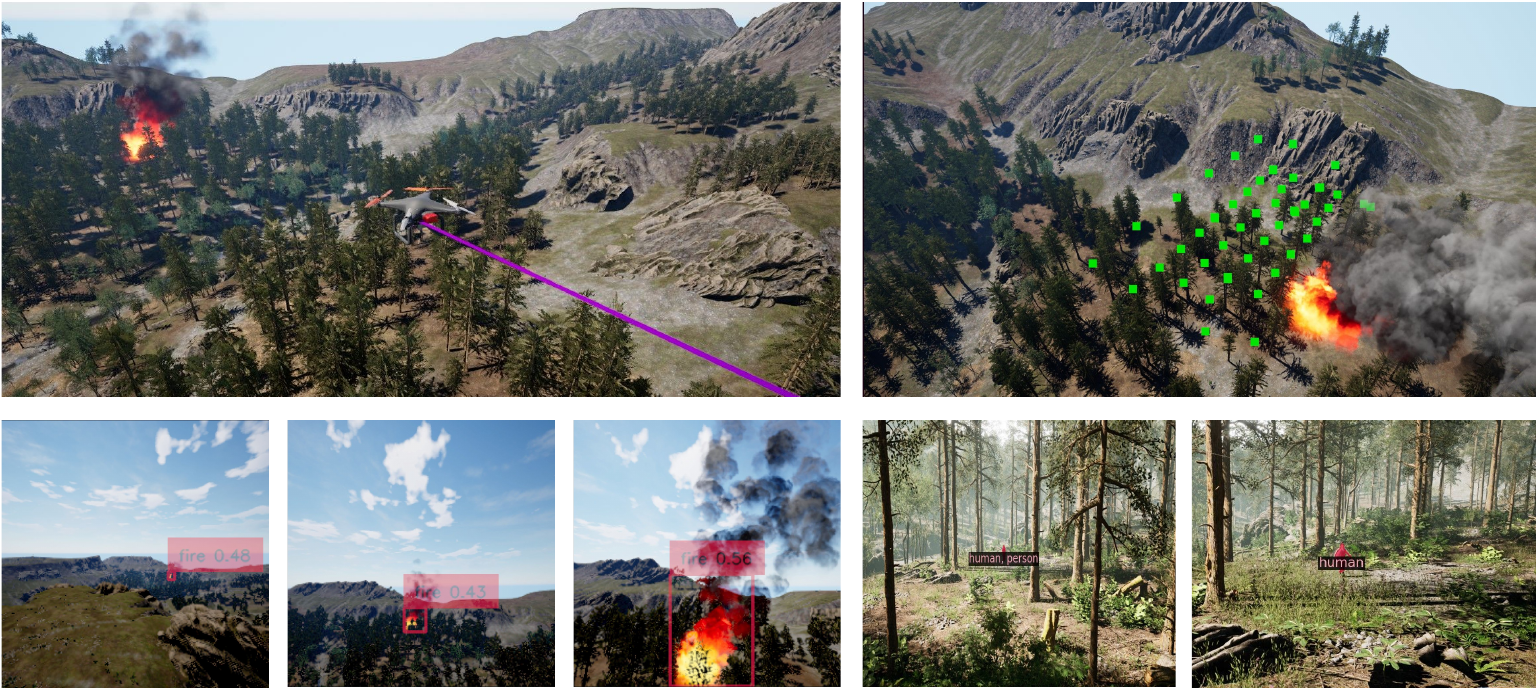}
     \caption{\small{\method{} can help solve complex robotics tasks, for example surveying areas affected by wild fires. (Top-left) With appropriate input \method{} is able to plan and execute flight and (top-right) given the observations from its sensor suite plan reconnaissance mission while avoiding the danger. (Bottom-row) Ability to invoke correct ML foundation models further helps solve difficult segmentation and detection tasks with ease.}}
    \label{fig:wildfire}
    \vspace{-3mm}
\end{figure}

\method{} provides a newer way of infusing intelligence into robots and we describe several experiments and scenarios to demonstrate how it accelerates engineering and lowers the barrier for roboticists and ML engineers. The current sets of experiments are demonstrated on an aerial robot platform within the AirGen simulator and we aspire to add other form factors soon. These experiments were carried out with the following set of foundation models as part of \method{}: GroundingDINO (object detection)\cite{liu2023grounding}, GroundedSAM (segmentation) \cite{kirillov2023segany}, MIDAS (RGB2Depth) \cite{ranftl2020midas}, DPVO (visual odometry)  \cite{teed2023deep}, TAPIR (point tracking) \cite{doersch2023tapir}, Optical Expansion (time-to-collision) \cite{yang2020upgrading}, LLaVa (visual Q\&A) \cite{liu2023llava}, GIT (visual Q\&A) \cite{wang2022git}, BLIP-2 (captioning) \cite{li2023blip2}. We use GPT-4 \cite{openai2023gpt4} as the LLM for reasoning and code generation, which when augmented with models classes such as VQA unlocks rich scene parsing abilities. We make several hypotheses and validate them using the GRID platform as follows.

\textbf{H1: GRID enables rapid prototyping of AI capabilities for robotics applications.}

We use the foundation models in GRID along with the large language model interaction layer to set up and solve a wildfire search and rescue scenario using a drone. The scenario assumes a potentially non-technical user interacting with GRID and thereby with a drone in the AirGen simulator. The drone is equipped with front and bottom facing cameras, and GRID is granted access to the control and sensor APIs of the drone.

We observe that this scenario which is a fairly complex combination of perception, control, and reasoning, can be prototyped with ease using GRID. In order to set up this task, we converse with GPT-4 through GRID. GPT-4 operates as an orchestrator and code generator, parsing user intent and writing the necessary code for each part of the task. In this mode of operation, we achieve the following:

\begin{enumerate}
    \item Take off and search for a fire in the vicinity of the drone.
    \item Use an object detector and visual servoing functionality to navigate to the proximity of the fire.
    \item Figure out a search area and multiple vantage points to search for humans.
    \item Fly into the forest and execute a search for humans, again using the object detector.
\end{enumerate}

Figure~\ref{fig:wildfire} shows snapshots of how \method{} performs in this scenario. Top-left highlights the trajectory that the quadrotor decides upon, while tracking the fire using an object detector, while top-right shows the grid search pattern the orchestration layer recommends given the perception inputs. The lower row indicates how \method{} is able to successfully pull relevant models for detection and segmentation of various entities of interest. 

Of particular note is the ability of the GRID to learn and store reusable skills. For example, the visual servoing functionality (tracking an object of interest and moving towards it), which is a widely used capability in robotics, was initially programmed through a conversation with GPT-4 where the user described the expected behavior. GPT-4 was asked to then write a textual description for itself on how visual servoing works, and in further experimentation, GPT was able to retrieve this skill and program it with ease. Similarly, we can also create modular reusable code blocks for different fundamental abilities.

\subsection{H2: Reasoning via LLMs can be useful for complex robotics scenarios}
\begin{figure}
    \centering
    \includegraphics[width=1.0\columnwidth]{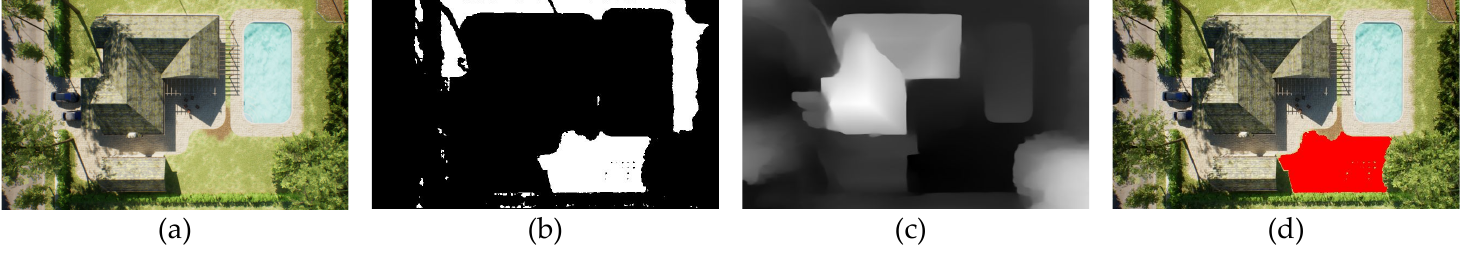}
    \caption{\small{GRID identifies a safe landing zone for a drone. Given an input RGB image (a), GRID first uses a VQA model which suggests the location of grassy areas. GRID then uses a segmentation model to segment out the grass (b), then an RGB to depth model to estimate the flatness of the terrain (c), which results in the final output (d).}}
    \label{fig:landingzone}
    \vspace{-3mm}
\end{figure}

To highlight the idea that language models can play a key role in robot intelligence, we set up a few problems that involve reasoning and orchestration of the foundation mosaic. We note that robotics often involves decision making not from a single model's point of view, but several. One example that highlights the need for multiple models is when we ask GRID to identify safe landing zones for a drone from a top down image (acquired from the AirGen simulator). \Cref{fig:landingzone} highlights how \method{} successfully addresses the problem. Having access to a multitude of domain specific expert models and a powerful orchestration layer allows GRID to solve this problem in a fairly complete fashion through the following steps. A particularly notable thing is that there was no user input in the entire process except for the initial request. The LLM comes up with the plan and calls all necessary models by generating the appropriate code.

\begin{enumerate}
\item GRID receives an image (\Cref{fig:landingzone}(a)) and invokes a vision-language model (LLaVa) to receive a holistic understanding of the scene. LLaVa identifies the existence of grassy areas that could allow safe landing.
\item GRID then leverages a segmentation model (GroundedSAM) to segment the grassy areas.
\item As a potential landing spot needs to be reasonably flat, GRID then knows to use an RGB to depth model (MiDaS) to get the depth map of the scene, and computes the intersection with the segmentation mask of the grassy area to find the flatter regions.
\item Finally, GRID identifies a safe, flat region for landing. 
\end{enumerate}

{\scriptsize	
\begin{mdframed}
\user{
\begingroup\normalfont
  \includegraphics[height=50px]{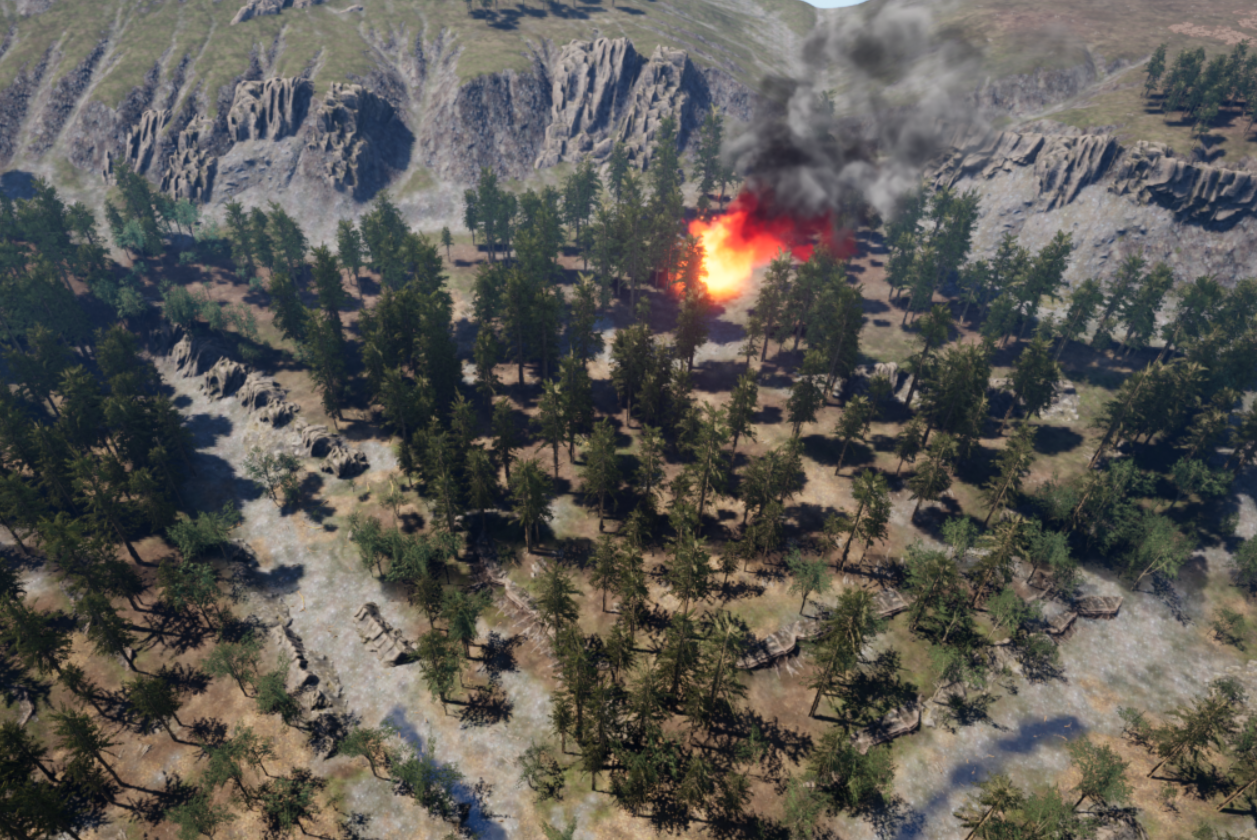}%
  \endgroup \\
I want you to analyze this image and identify some good locations for a search and rescue operation.}\\
\llm{To answer this question, we would need to make several assumptions and considerations. Here are some of them:
1. Survivors are likely to be found in areas that are not directly affected by the fire. 
2. They could be found in relatively open and safe areas where they might have attempted to escape the fire.
3. They could be in areas close to buildings or structures, as people are generally found near such locations.
4. It's unlikely to find survivors in areas with dense smoke or fire.\\
Given these considerations, here's a potential plan:\\
1. Read the image.\\
2. Use the LLaVA model to get a general sense of the scene. Ask it general questions about the scene to get an idea of what's in it.\\
3. Use the GroundingDINO model to detect areas of fire and smoke. We can then exclude these areas from the heatmap of probable locations for survivors.\\
4. If there are visible structures or buildings, use the GroundingDINO model to detect them, and include these areas in the heatmap.\\
5. Generate a heatmap based on the above factors indicating the areas where it's most likely to find survivors.\\
Let's try to confirm this by asking the model directly about the presence of fire in the image.\\
\codebox{GPT call to LLaVa}\\
The VQA model now identifies a forest with trees and a fire burning in the background. The fire is creating smoke and appears to be in a wooded area.\\
Let's start by asking the VQA model about potential safe areas in the image.\\
\codebox{GPT call to LLaVa} \\
The VQA model suggests several potential safe locations for survivors: \\
1. Forest with trees providing shelter and cover from the fire and smoke.\\
2. A river nearby which can offer a source of water and a means of escape.\\
3. A rocky cliff in the background that could provide temporary refuge, although it is noted that its proximity to the fire and the smoke may not be ideal for long-term safety.\\
To accurately mark these locations, we need to determine the rough coordinates of the mentioned points. We can use the object detection model to locate these areas.\\
... \textit{conversation continues}
}
\end{mdframed}
}

{\scriptsize	
\begin{mdframed}
\user{
\begingroup\normalfont
  \includegraphics[height=50px]{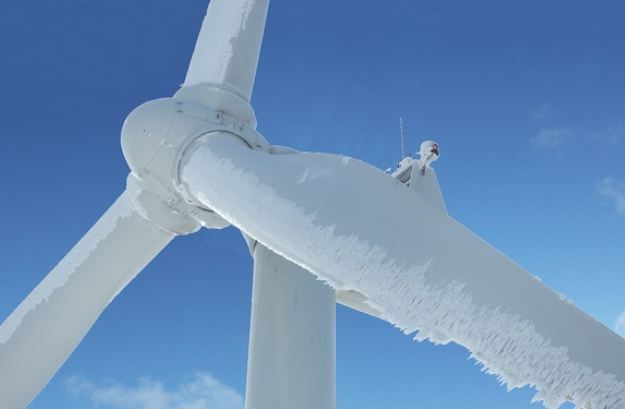}%
  \endgroup \\
Analyze the turbine blades in this image.}\\
\llm{The turbine blades are covered in snow, which indicates that the turbine is located in a cold and snowy environment. The snow accumulation on the blades could potentially affect the turbine's performance, as it may add weight and reduce the efficiency of the turbine. }\\
\user{
\begingroup\normalfont
  \includegraphics[height=50px]{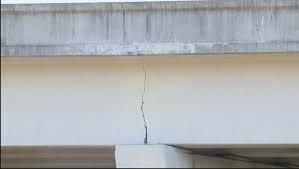}%
  \endgroup \\
Analyze the structure in this image.}\\
\llm{The image shows a concrete bridge with a damaged section, where the concrete has cracked and fallen away. The bridge appears to be in a state of disrepair, with the damaged area being exposed to the elements. This type of damage can pose a risk to the structural integrity of the bridge and may require immediate repairs to prevent further deterioration or potential collapse. }
\end{mdframed}
}

In a similar fashion, when asked to identify potential vantage points to look for survivors in a forest fire scene, GRID harnesses the combined abilities of the vision-language assistant and an object detection model to identify possibilities. The models within GRID are also capable of transforming imagery into coherent natural language descriptions. This makes them invaluable for applications like drone-based inspections, where immediate and actionable feedback is crucial, and they can help streamline decision-making processes by providing clear insights derived from vast amounts of data.

\textbf{H3: Foundation Mosaic can address zero-shot task completion}

\begin{figure}
    \centering
    \includegraphics[width=\linewidth]{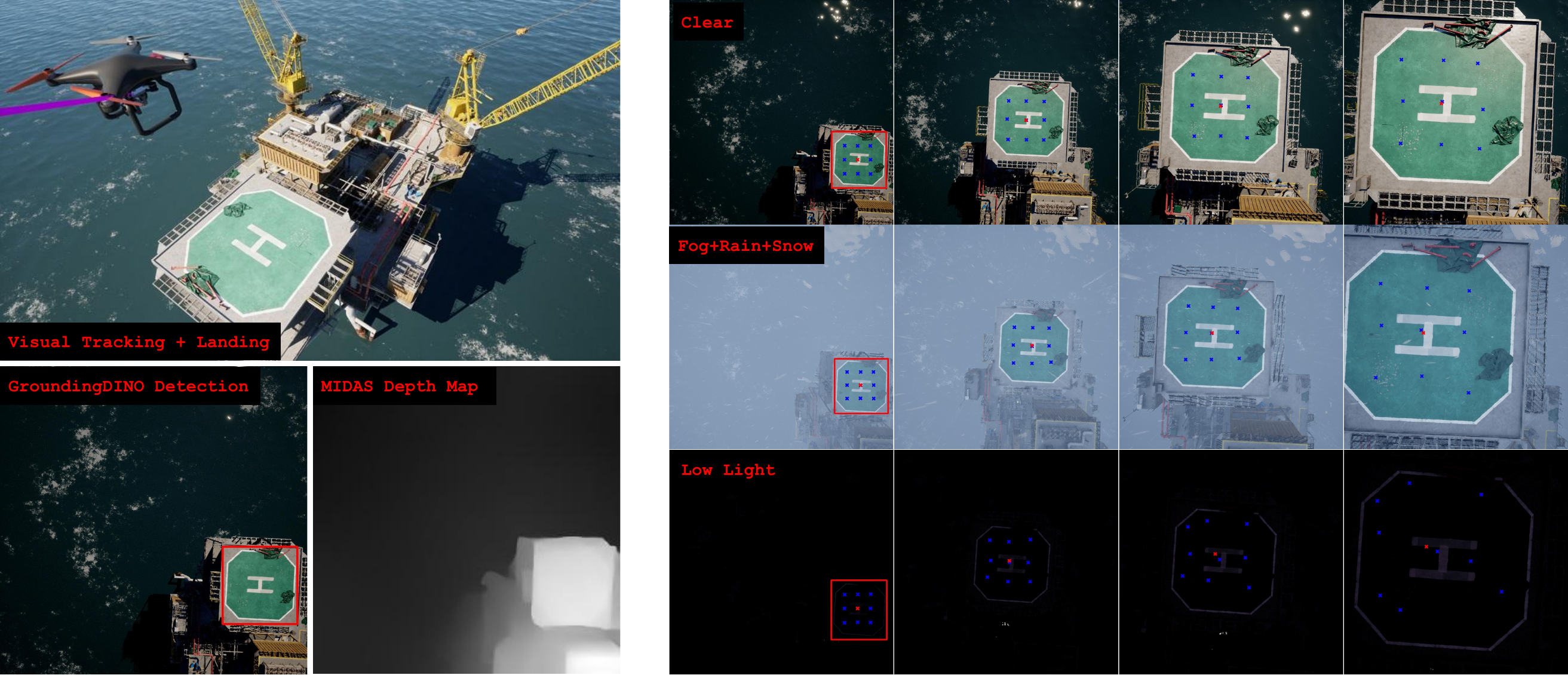}
    \caption{\small{\method{} can be used to orchestrate models for a vision based autonomous landing task. To land the drone, GroundingDINO \cite{liu2023grounding} is used to detect the helipad and then TapNet \cite{doersch2023tapir} tracks a grid of points near its center. The drone uses MiDaS \cite{ranftl2020midas} for monocular depth estimation which helps estimate the target velocity during landing. The drone is able to land even in adverse weather conditions (snow/fog/low light).}}
    \label{fig:oilrig}
    \vspace{-3mm}
\end{figure}

The fact that the foundation mosaic is composed of several models, and is designed with generalizability as the primary focus offers a much lower sample complexity when solving complex robotics tasks. We set up a vision based autonomous landing task for a drone in a synthetic environment (which mimics an oil rig). The aim of the task is to build a perception-action loop so that the drone can be automatically guided to a proper landing spot with downward-facing cameras. \Cref{fig:oilrig} highlights the results. We orchestrate the landing task through a combination of models within GRID: we use GroundingDINO \cite{liu2023grounding}, an open-set object detector to first detect the landing pad, from which we extract a few salient points for tracking. We use Tapnet \cite{doersch2023tapir} to track these points over the course of time from the stream of images, while simultaneously using MiDaS to estimate depth from monocular imagery to estimate target velocity. Through this combination of pretrained models, we observe that the visual landing module works out of the box even under adverse weather conditions (simulated snow and fog), as well as low-light conditions. 

\textbf{H4: Foundation Mosaics enable safety even with limited sensing}

\begin{figure}
    \centering
    \includegraphics[width=1.0\columnwidth]{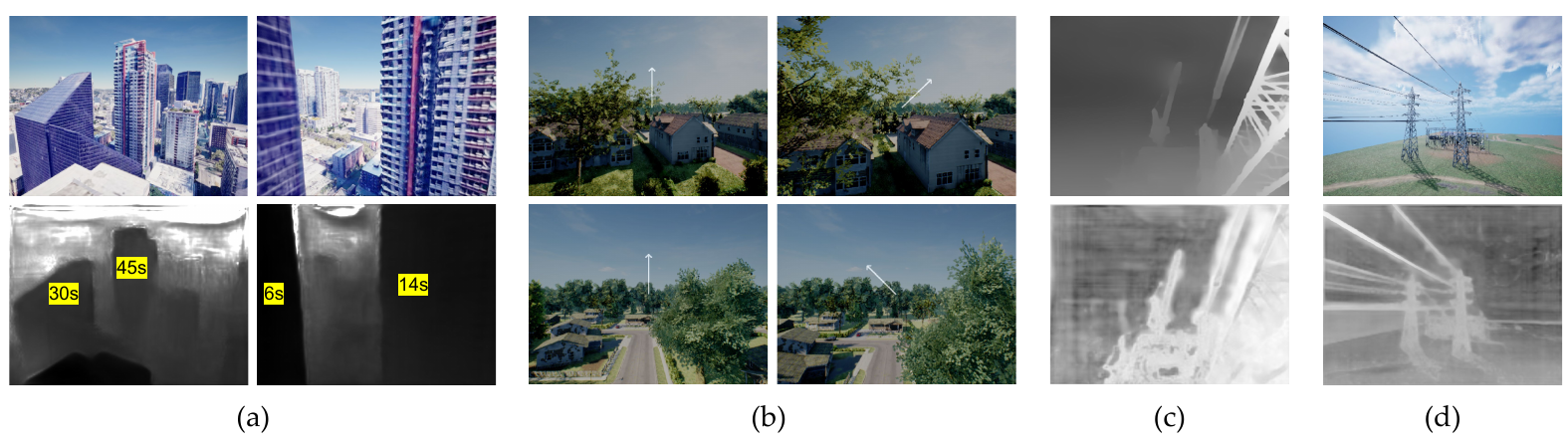}
    \caption{\small{The GRID foundation models help extract a multitude of features from RGB images that contribute to safe navigation. (a) An optical expansion based model estimates time to collision (TTC) during flight from RGB imagery. (b) Inferred safe velocity vectors from the RGB imagery (c) Depth maps and occupancy predictions from RGB imagery, (d) Occupancy predictions also capture tricky objects such as thin wires.}}
    \label{fig:safety}
    \vspace{-3mm}
\end{figure}

The combination of foundation models in GRID allows for extracting a holistic understanding of the world even from limited sensing. We examine the case of a drone equipped only with an RGB camera, and rely on a combination of models to extract depth from monocular data (MIDAS), as well as motion in depth through optical expansion \cite{yang2020upgrading}. The latter is also converted into a time to collision map, which further allows for safe navigation and estimation of collision free actions. Similar to the earlier examples with GPT-4 making use of foundation models, we prototype a quick obstacle avoidance algorithm using GPT-4 and the TTC model. Figure~\ref{fig:safety} shows the results where from RGB images we can infer safe sets of actions as well as construct depth and occupancy maps. \Cref{fig:safety}(b), where the arrows indicate a predicted safe action by the algorithm written by GPT whereas Figure~\ref{fig:safety}(c) and (d) show prediction of occupancy maps. 

\textbf{H5: \method{} is useful for data generation}

Critical to applications such as machine learning training, as well as real life scenarios such as regular inspection and maintenance is the ability to generate data. Often, this takes the form of being able to record visual or other modalities of data while observing a certain scene, or an object. We use GRID for an infrastructure inspection scenario in AirGen for a wind turbine. Given an RGB-D image of the object, the task is to extract a trajectory that scans the object to the fullest extent by flying close to its surface. 

\begin{figure}
    \centering
    \includegraphics[width=0.7\columnwidth]{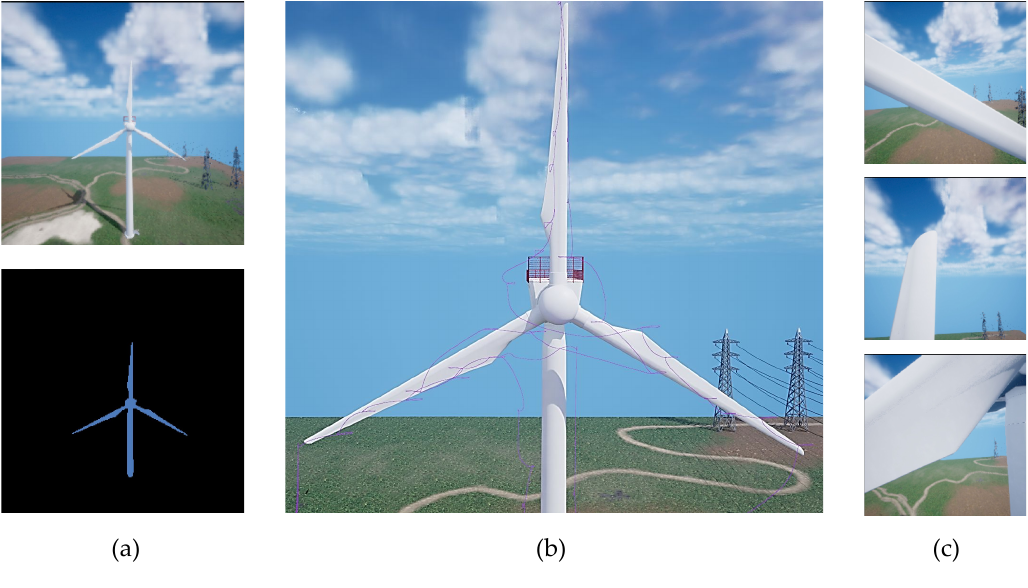}
    \caption{\small{\method{} provides a rich framework to generate data by combining ML constructs with classical robotic primitives. (a) Upon observing an asset that needs inspection, \method{} is able to compute segmentation masks with appropriate model, (b) plan a trajectory (shown in red) via computing surface normals and then invoking an optimal path planner and (c) generate images as it traverses the trajectory.}}
    \label{fig:turbine}
    \vspace{-3mm}
\end{figure}

To achieve object inspection, we combine the foundation models with computer vision and planning primitives. Figure~\ref{fig:turbine} highlights the observations. Given an RGB image, the inspection module first computes a segmentation mask for the object of interest using one of the segmentation models in GRID (figure~\ref{fig:turbine}(a)). For the turbine example scenario, we use GroundedSAM \cite{liu2023grounding, kirillov2023segany}. Once the segmentation mask is computed, we use computer vision primitives to extract the point cloud from the depth image, from which we compute surface normals. Once the surface normals are available, these are translated into potential inspection points for the drone, and we use an optimal path planner to connect these points for an inspection trajectory. Figure~\ref{fig:turbine}(b) shows the automatically planned path in red and highlights the power of the \method{} platform where state-of-the-art ML methods and models are seamlessly being combined with robotics primitives. Figure~\ref{fig:turbine} shows sample images that were collected as part of the inspection routine.

Given that GRID combines simulation with foundation models and computer vision libraries, we envisage an effective scenario for real life automated inspection / data generation. A real object mesh can be imported into the simulation (directly from the ground truth mesh if one is available, or through the NeRF class of techniques), upon which the inspection module can be run, leveraging the rich modalities in simulation such as depth, surface normals etc. available as ground truth. Once an inspection trajectory is computed, this can be evaluated and modified in the simulation, potentially with the language model acting as an interface, and finally converted into waypoints for a real drone and flown in real life.

\section{Related Work}
\textbf{Composing Pretrained Models with LLMs:}
There is a wave of recent work on composing/orchestrating a set of tools (external pretrained models, library functions, pieces of code etc.) both by fine-tuning and using the \textit{in-context learning} ability of LLMs.
Toolformer \cite{schick2023toolformer} showed how generating a dataset of \textit{(instruction, API)} pairs to fine tune an LLM can equip it with basic abilities of being able to use those tools in the appropriate contexts. Gorilla \cite{patil2023gorilla} is a LLaMA-based model fine-tuned on a more diverse dataset of \textit{(instruction, API)} pairs that is able to convert natural language instructions to code. Socratic Models \cite{zeng2022socratic} used a composition of visual and text models with a timestamped world state text history to perform zero-shot reasoning tasks.
VisProg \cite{gupta2023visual} and ViperGPT \cite{suris2023vipergpt} demonstrate compositional visual reasoning with text inputs by generating python code from text instructions and available models. TaskMatrix \cite{liang2023taskmatrix}, Chameleon \cite{lu2023chameleon}, HuggingGPT \cite{shen2023hugginggpt} and Prismer \cite{liu2023prismer} are similar ideas that compose existing pretrained vision and language models. 
Hsieh et al. \cite{hsieh2023tool} demonstrate the effectiveness of using API documentation instead of in-context examples of API usage for composing a set of tools. Yang et al. \cite{yang2023foundation} discuss various methods and challenges in using Foundation Models for real-world decision making scenarios.
Foundation Mosaics in GRID utilize a combination of these orchestration techniques (chain-of-thought prompting, in-context examples, model, robot, and sensor API descriptions) to enable a robot to perform a complex task using natural language instruction. Our implementation of Foundation Mosaics enables LLMs to orchestrate domain and robot-specific modules, serve as explainable reasoning agents, and act as intuitive interfaces for human-machine interaction across a diverse variety of robotics scenarios.

\textbf{Learning robotics representations:}
There has been a lot of interest in representation learning and many of them are relevant to robotics. Examples include
SayCan \cite{ahn2022can}, PACT \cite{bonatti2022pact}, RT1 \cite{brohan2022rt}, RT2 \cite{brohan2023rt}, Gato\cite{reed2022generalist}, Palm-E \cite{driess2023palm}, RoboAgent \cite{bharadhwaj2023roboagent}, VC-1\cite{majumdar2020improving}, R3M \cite{nair2022r3m}, VIP \cite{ma2022vip}, COMPASS \cite{ma2022compass}, PVR for control \cite{parisi2022unsurprising}, masked visual pretraining \cite{radosavovic2023real}, offline visual representation learning for embodied AI \cite{yadav2023offline}. Approaches for task learning in robotics can be categorized into three main segments. The first, and the most common, is the task-specific training approach where modules are designed specific to each task. Such methods have been proposed for several of the most common robotic tasks such as visual/LiDAR based localization, mapping, path planning and control ~\cite{li2019net, barsan2020learning}. Another category involves multi-task learning approaches, where models are trained jointly to be able to solve several tasks~\cite{kalashnikov2021mt, huang2021generalization}. Finally, there exists a class of techniques that perform task-agnostic pre-training, whose representations can later be finetuned for a task of choice~\cite{parisi2022unsurprising,ma2022compass,nair2022r3m}.

\textbf{Robotics datasets:} There also have been efforts to curate dataset for robotics and autonomous systems and include a number of scenarios such as aerial \cite{tartanair2020iros}, manipulation \cite{zhu2020robosuite}, embodied agents \cite{grauman2022ego4d, bahl2023affordances}, autonomous driving \cite{waymo} and with multiple form factors \cite{brohan2022rt, majumdar2023we, li2023internet}.

\textbf{Multi-modal robotics representations:}
Robotics and Autonomous systems require rich, well-grounded representations and benefit from the existence of multiple sensors of different modalities. Robotics tasks such as manipulation have been shown to benefit from object-centric representations \cite{devin2018deep, florence2018dense}, and combining geometry has been shown to be effective for navigation tasks. Multimodal representation learning has been applied to robotic manipulation and grasping in the form of visuo-tactile representations \cite{lee2019making}, as well as scene understanding and exploration by combining RGB and depth \cite{chaplot2020object}, and human robot interaction \cite{campbell2019probabilistic}. Cross-modal representation learning has been combined with imitation learning to result in drone navigation policies \cite{bonatti:2020:cmvae}. Multi-task learning has been examined for self-driving under different modes \cite{chowdhuri2019multinet} and visual odometry/semantic segmentation \cite{radwan2018vlocnet++}. 

Learning multimodal representations has been examined in several domains such as vision-language \cite{li2020unicoder,ahuja2020no}, vision-audio \cite{ginosar2019learning,owens2018audio}, image registration \cite{roche1998correlation,hu2018weakly}, and video understanding \cite{sun2019learning, gordon2020watching}. Tsai et al.~\cite{tsai2018learning} present a framework learning intra-modal and cross-modal interactions from input, and Alayrac et al.~\cite{alayrac2020self} present a multimodal learning approach for text, audio and video. 
Inspired by the success of large-scale pretraining in the text domain \cite{devlin2018bert, brown2020GPT-3}, pretrained models have also been developed for vision-language tasks \cite{lu2019vilbert, zhang2021vinvl, li2020oscar}. A natural extension of multimodal learning algorithms has been applied to the multi-task learning setting \cite{pramanik2019omninet, kuga2017multi, chaplot2019embodied}. Numerous surveys on multimodal learning are also largely focused on vision, text and speech~\cite{ramachandram2017deep,zhang2020multimodal,guo2019deep}. Baltrusaitis et al.~\cite{baltruvsaitis2018multimodal} point out the opportunity for co-learning with multimodal data where knowledge from one (resource rich) modality can be exploited in modeling another (resource poor) modality.  

Visual-language navigation of embodied agents is well-established field with clear benchmarks and simulators~\cite{szot2021habitat,Anderson2018room2room}, and multiple works explore the alignment of vision and language data by combining pre-trained models with fine-tuning~\cite{hao2020genericVLN,thomason2020vision,nguyen2019helpAnna}
To better model the visual-language alignment, \cite{ma2019progressestimation} also proposed a co-grounding attention mechanism.
In the manipulation domain we also find the work of \cite{shridhar2022cliport}, which uses CLIP~\cite{radford2021learning} embeddings to combine semantic and spatial information.

The existing visual-language representation approaches primarily rely on BERT-style~\cite{devlin2018bert} training objectives to model the cross-modal alignments. 
Common downstream tasks consist of visual question-answering, grounding, retrieval and captioning etc.~\cite{sun2019videobert, lu2019vilbert, zhou2020unified, su2019vlbert}. Learning representations for robotics tasks poses additional challenges, as perception data is conditioned on the motion policy and model dynamics~\cite{bommasani2021opportunities}.

\textbf{Transformers in robotics:}
Transformers were originally introduced in the language processing domain~\cite{vaswani2017attention}, but quickly proved to be useful in modeling long-range data dependencies other domains.
Within robotics we see the first transformers architectures being used for trajectory forecasting~\cite{giuliari2021transformer}, motion planning~\cite{bucker2022reshaping,chaplot2021differentiable}, and reinforcement learning~\cite{chen2021decision,janner2021offline}.
The main difference between these works and \method{} is that they are focused on training a model for a single task, while we propose learning representations amenable to multiple downstream tasks for a robot.

\textbf{Simulation:}
Inspired by the emergence of pretrained representations as a powerful abstraction in computer vision, natural language, and speech, recent studies have been trying to leverage simulators, like MuJoCo~\cite{todorov2012mujoco}, Habitat~\cite{savva2019habitat}, Gibson~\cite{xia2018gibson} in order to generate embodied / egocentric data to capture relevant traits for control tasks.  Some evidence was shown that such visual-motor representations can be useful for downstream control tasks, specially lower-level features (early layers in feature hierarchy) that encode spatial information~\cite{parisi2022unsurprising}, while higher-level features are more useful when a semantic understanding of the environment is needed (e.g. in Habitat for instruction following or ObjectNav). One difficulty on using many of these environment is the fact that most of them are still in development and do not offer all the tooling needed for data generation or evaluation for fine-tuning tasks in the robotic domain like localization or mapping. Additionally, most of the current simulators are one-robot or one-task toy environments, which are not realistic for robotics applications. Usually a robot needs to perform multiple secondary objectives (mapping, localization, planning, object identification) to perform a complex task well. We need to involve robotics with the lenses of a neural fabric if we want to achieve more realistic applications.

\section{Conclusion and Future Work}
\method{} provides a novel way to infuse intelligence into robots and autonomous systems. The key idea is to build upon recent advances in representation learning and foundation models and provide a framework that can significantly reduce the effort for roboticists and ML engineers. Ability to orchestrate between various representations, robot foundation models, simulation and real-world operations enables ML mechanisms, such as simulation feedback, that are unique and will lead to lowering the barrier to entry into robotics. 

\method{} is an evolving platform and we hope to incorporate several form factors, including robotic arms, ground vehicles, underwater robots among others in future. There is much that needs to be explored in the real of newer mechanisms such simulation feedback and agent based architectures. As the platform matures we expect to curate large amounts of multimodal robot data fueling significantly larger foundation models. Safety is an important consideration that needs further research. Also it will be important to investigate optimal scaling characteristics for multimodal robotic models.

\footnotesize{
\printbibliography
}

\newpage

\end{document}